%% file: main.tex
\documentclass[10pt,twocolumn,letterpaper]{article}

\usepackage{cvpr}
\usepackage{times}
\usepackage{epsfig}
\usepackage{graphicx}
\usepackage{amsmath}
\usepackage{amssymb}
\usepackage{microtype}
\usepackage{booktabs}
\usepackage{tabularx}

\usepackage[breaklinks=true,bookmarks=false]{hyperref}

\cvprfinalcopy 

\def\cvprPaperID{****} 

\begin{document}

\title{Shallow and Deep Convolutional Networks for Saliency Prediction}

\author{Junting Pan\thanks{Equal contribution.} ,\ Elisa Sayrol and Xavier Giro-i-Nieto\\
Image Processing Group\\
Universitat Politecnica de Catalunya \\
Barcelona, Catalonia/Spain\\
{\tt\small xavier.giro@upc.edu}
\and
Kevin McGuinness\footnotemark[1] \ and Noel O'Connor\\
Insight Center for Data Analytics\\
Dublin City University\\
Dublin, Ireland\\
{\tt\small kevin.mcguinness@insight-centre.org}
}

\maketitle

\input{0_abstract}

\input{1_motivation}
\input{2_related_work}
\input{3_juntingnet}
\input{4_salnet}

\input{5_results}
\input{6_conclusions}
\input{9_acknowledgements}

{\small
\bibliographystyle{ieee}
\bibliography{egbib}
}

\end{document}



\title{ Supplementary material for Paper ID 1843}
\maketitle


In the following table the results when fine-tunning with MIT300 are shown.

Comparing the results of this table with table 6 we see an improvement ....


%% file: 0_abstract.tex
\begin{abstract}
The prediction of salient areas in images has been traditionally addressed with hand-crafted features based on neuroscience principles. This paper, however, addresses the problem with a completely data-driven approach by training a convolutional neural network (convnet). The learning process is formulated as a minimization of a loss function that measures the Euclidean distance of the predicted saliency map with the provided ground truth. The recent publication of large datasets of saliency prediction has provided enough data to train end-to-end architectures that are both fast and accurate. Two designs are proposed: a shallow convnet trained from scratch, and a another deeper solution whose first three layers are adapted from another network trained for classification.
To the authors knowledge, these are the first end-to-end CNNs trained and tested for the purpose of saliency prediction.
\end{abstract}

%% file: 1_motivation.tex
\section{Introduction}

\begin{figure}%
		\includegraphics[width=\linewidth]{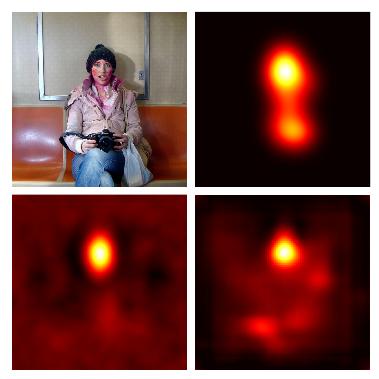}
		\caption{Input Image (top left) and saliency maps from the ground truth (top right), our shallow convnet (bottom left) and our  deep convnet (bottom right).}
		\label{fig:examples}
\end{figure}

This work presents two approaches of end-to-end convolutional neural networks (convnets or CNNs) for saliency prediction. 
Our objective is to compute saliency maps that represent the probability of visual attention on an image, defined as the eye gaze fixation points. 
This problem has been traditionally addressed with hand-crafted features inspired by neurology studies.
In our case we have adopted a completely data-driven approach, using a large amount of annotated data for saliency prediction.
Figure \ref{fig:examples} provides an example of an image together with its ground truth saliency map and the two saliency maps predicted by the proposed convnets: a shallow one and a deep one.

Convnets are popular architectures in the field of deep learning and have been widely explored for visual pattern recognition, ranging from a global scale image classification ~\cite{krizhevsky2012imagenet} to a more local object detection ~\cite{girshick2014rich} or semantic segmentation ~\cite{long2015fully}.
The hierarchy of layers in convnets are inspired by biological models, and some works have pointed at a relation between the activity of certain areas in the brain with hierarchy of layers in the convnets ~\cite{agrawal2014pixels, cichy2015mapping}.
Provided with enough training data, convnets have shown impressive results, often outperforming other hand-crafted methods. 
The rise of convnets started from the computer vision task where more annotated data can be easily collected, that is, image classification~\cite{russakovsky2015best}.
Large datasets like ImageNet~\cite{imagenet_cvpr09} or Places~\cite{zhou2014learning} have provided enough visual examples to train the millions of parameters that most popular convnets contain.
These datasets provide thousands of images for each discrete label  typically associated to a semantic class.

The saliency prediction problem, however, poses two specific challenges different from the classic image classification.
First, collecting large amount of training data is much more costly because it requires capturing the fixation points of human observers instead of a textual label for each image.
Our work has benefited from recent publications of two 
large datasets containing images and an annotation of their salient points for humans \cite{jiang2015salicon, xu2015turkergaze}.
%
Collecting this level of of data has been possible thanks to crowdsourcing approaches, the same strategy used to annotate the ImageNet and Places datasets.

The second challenge to address when using convnets for saliency prediction is that a saliency score must be estimated for each pixel in the input image, instead of a global-scale label for the whole image.
The saliency map at the output must present a spatial coherence and a smooth transition between neighbouring pixels.

The main contribution of this work is addressing the saliency prediction problem from an end-to-end perspective, by using convnets for regression rather than classification. 
We apply this strategy with two different architectures trained with two different approaches:
a shallow convnet trained from scratch, and a deep convnet that reuses parameters from the bottom three layer of a network previously trained for classification.
To the authors knowledge, these were the first convnets that formulate saliency prediction as an end-to-end regression problem.





This paper is structured as follows. Section~\ref{sec:RelatedWork} presents the previous and recent works using convolutional networks for saliency prediction and detection.
Section~\ref{sec:JuntingNet} introduces the shallow convnet, while Section~\ref{sec:SalNet} presents the deep network.
Section~\ref{sec:experiments} compares both networks in terms of memory requirements. It also shows, prediction performance in the MIT Saliency Benchmark and LSUN Saliency Prediction Challenge 2015 and they are compared with other models.
Conclusions and future directions are outlined in Section~\ref{sec:conclusions}.

Our results can be reproduced with the source code and trained models available at \url{https://github.com/imatge-upc/saliency-2016-cvpr}.

%% file: 2_related_work.tex
\section{Related work}
\label{sec:RelatedWork}

The proposed networks presents the next natural step to two main trends in deep learning: using convolutional neural networks for saliency prediction and training these networks by formulating saliency prediction as an end-to-end regression problem.
This section reviews related work in these directions.


An early attempt of predicting saliency with a convnet was the \textit{ensembles of Deep Networks (eDN)}~\cite{vig2014large}, which proposed an optimal blend of feature maps from three different convnet layers, that were finally combined with a simple linear classifier trained with positive (salient) or negative (non-salient) local regions.
This approach inspired \textit{DeepGaze}~\cite{kummerer2014deep} to adopt a deeper network. In particular, \textit{DeepGaze} used the existing AlexNet network~\cite{krizhevsky2012imagenet}, where the fully connected layers were removed to keep the feature maps from the convolutional layers. The response of each layer were fed into a linear model and its weights learned. 
DeepGaze would be the first case of transfer learning from a convnet for classification used for saliency, as we propose in our deeper architecture. However, we do not train a linear model to combine feature maps but directly train a stack of new convolutional layers on top of the transferred ones.
Other recent works have explored the combination of different convnets working at different resolutions to capture both global and local saliency.
Liu \etal \cite{liu2015predicting} proposed an architecture with three convnets working in parallel
where the three final fully connected layers are combined in a single layer to obtain the saliency map. Unlike our work the network is trained with image regions centered on fixation and non-fixation eye locations. On the other hand, a related preprint by Srinivas and Ayush~\cite{srinivas2015deepfix} appeared during submission of our work. Their model captures information at different scales by using very deep networks. Their work is inspired by the VGG network architecture proposed by Simonyan and Zisserman~\cite{simonyan2015verydeep}. Very deep networks may obtain richer information of image semantics. Some layers use inception style convolutional blocks~\cite{Szegedy2015goingdeeper} that capture semantics at different scales. By using large receptive fields, global context is also incorporated. As in the  \textit{DeepGaze} proposal, the network needs to be trained with databases that are not specific for eye fixation but are useful to capture generic image features.

Other approaches introduce new architectures and improvements in salient object detection. Zhao \etal \cite{zhao2015saliency} use also two parallel networks to obtain local and global context modeling. The input image consists of a superpixel-centered window that is preprocessed differently to feed each of the two convnets. Fully connected layers are combined at the end to obtain the salient objects. The work by Li and Yu~\cite{liu2015visual} proposes three nested windows as inputs to three different convnet at different scales that are fused together to obtain an aggregated saliency map. Wang \etal. proposed a different pipeline~\cite{wang2015deepnetworks}: local estimation is carried out and the resulting information is used as input to obtain a global search. That is, first, to detect local saliency, a deep neural network (DNN-L) learns local patch features to determine the saliency value of each pixel. Second, the local saliency map together with global contrast and geometric information are used as global features to obtain object candidate regions. A deep neural network (DNN-G) is then trained to predict the saliency score of each object region based on global features. Finally, a very recent work introduced by Li \etal \cite{li2015deepsaliency}, combines semantic image segmentation and saliency detection, sharing the first layers that exploit extraction of effective features for object perception. Only the last layers are divided to obtain the corresponding segmentation and saliency detection images. This proposal is also inspired by the VGG very deep network introduced in~\cite{simonyan2015verydeep}. In this case, 15 layers and pretraining with the Imagenet dataset is used. 


Fully Convolutional Networks (FCNs)~\cite{long2015fully} addressed the semantic segmentation task to predict the semantic label of every individual pixel in the image. 
This approach dramatically improved previous results on the challenging PASCAL VOC segmentation benchmark~\cite{everingham2014pascal}.
The idea of an end-to-end solution for a 2D problem as as semantic segmentation was refined by \text{DeepLab-CRF}~\cite{DBLP:journals/corr/ChenPKMY14}, where the spatial consistency of the predicted labels is improved using a Conditional Random Field (CRF), similarly to the hierarchical consistency enforced in~\cite{farabet2013learning}.

In our work we are interested in finding saliency maps rather than salient object detection by training convnets end-to-end. We also focus on novel databases that are annotated for the purpose of saliency prediction.



%% file: 3_juntingnet.tex
\section{Shallow Convnet}
\label{sec:JuntingNet}

This section presents the first of our proposed convnets, which is based on a lightweight architecture whose parameters are trained from scratch.



\subsection{Architecture}
\label{ssec:juntingnet-arch}

\begin{figure}
  \centering
  \includegraphics[width=\columnwidth]{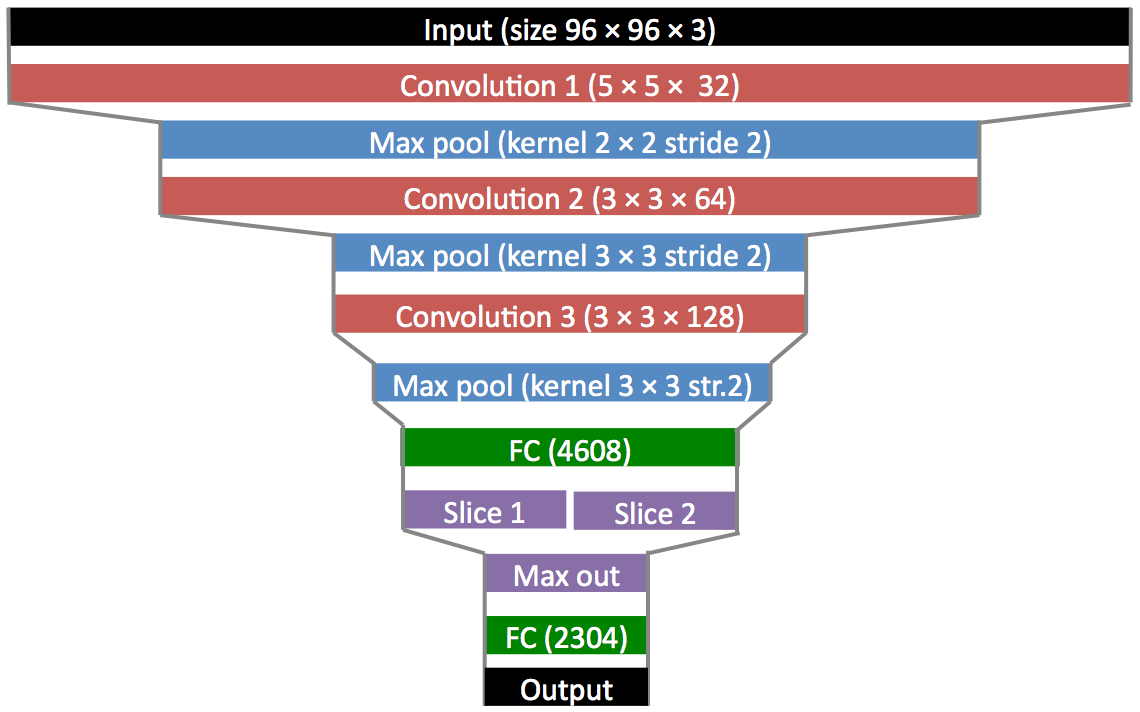}
  \caption{Architecture of the shallow convolutional network.}
  \label{fig:juntingnet}
\end{figure}

The network consists of five layers with learned weights: three convolutional layers and two fully connected layers.
Each of the three convolutional layers is followed by a rectified linear unit non-lineraity (ReLU) and a max pooling layers.
Figure \ref{fig:juntingnet} shows a detailed description of each layer. The network has to a total of 64.4 million free parameters. 

The network was designed considering the a amount of available saliency maps for training it from scratch.
Different strategies were considered to avoid overfitting the model.
First, we used three convolutional layers rather than the five used in the classic AlexNet architecture~\cite{krizhevsky2012imagenet} (and far less than very deep networks used recently such as the thirteen used in VGG-16 \cite{simonyan2014very}).
Second, the input images are resized to $[96 \times 96]$, a much smaller dimension that the $[227 \times 227]$ used in AlexNet~\cite{krizhevsky2012imagenet}.
The three max pooling layers reduce the initial $[96 \times 96]$ feature maps down to $[10 \times 10]$ by the last of the three  poolings.

Even with the above constraints, the network still overfits significantly. We found that norm constraint regularization for the maxout layers~\cite{goodfellow2013maxout}, which computes the \textit{max} between pairs of of the previous layer’s output,
was essential to mitigate this overfitting.
We also tested using dropout~\cite{hinton2012improving} after the first fully connected layer, with a dropout ratio of 0.5 (50\% of probability to set a neuron's output value to zero), but this did not improve overfitting much, and so was not included in the final model.

Notice that the 2,304-dimensional vector at the output is mapped into a 2D array of $[48 \times 48]$, which correspond to the saliency map.
This decrease in resolution is compensated at test time by resizing the dimensions of the output to match the input image and posterior filtering using a Gaussian kernel with a standard deviation of $2.0$.

This shallow convnet was implemented using Python, NumPy, and the deep learning library Theano~\cite{bergstra2010theano, bastien2012theano}. 
Processing was performed on an NVIDIA GTX 980 GPU with 2048 CUDA cores and 4GB of RAM. 
It took between 6 and 7 hours to train for the SALICON dataset, and 5 to 6 hours for the iSUN dataset.
Saliency prediction requires 200 ms per image.

\subsection{Training}

This shallow network was trained from scratch twice, each time from a different dataset. 
A first model was built using the $10,000$ saliency maps from the SALICON dataset \cite{jiang2015salicon}, and a second model using the $6,000$ saliency maps from the iSUN dataset.
Both datasets are described in detail in Section~\ref{ssec:datasets}.
Given the smaller amount of images available in the iSUN dataset \cite{xu2015turkergaze}, a slight modification was introduced in this second model:
the depth of the third convolutional network was of 64 instead of 128, as depicted in Figure~\ref{fig:juntingnet}.

The weights in all layers are initialized from a normal Gaussian distribution with zero mean and a standard deviation of 0.01, with biases initialized to 0.1.
The network was trained with stochastic gradient descent (SGD) and the Nesterov momentum method, which we found helps convergence. 
The learning rate changed over time, starting with a higher learning rate 0.03 and decreased during training to 0.0001. 
We trained the network for 1,000 epochs.
For validation purposes, we split the training data into 80\% for training and the rest for periodic validation. 
A data augmentation technique was used by mirroring all images. All considered saliency maps were normalized to $[0,1]$.

\begin{figure}
  \centering
  \includegraphics[width=0.3\columnwidth]{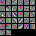}
  \caption{Filters learned for the first convolutional layer of the shallow convnet (best viewed from a distance).}
  \label{fig:juntingnet-filters}
\end{figure}

The filters learned in the first convolutional layer are shown in Figure \ref{fig:juntingnet-filters}.
They present a similar pattern to other similar filters learned for classification convnets \cite{zeiler2014visualizing, zeiler2014visualizing}, where edge detectors can be identified.
It is noticeable how these type of filters arise also when training our network on saliency maps.

%% file: 4_salnet.tex
\section{Deep Convnet}
\label{sec:SalNet}

The second approach explored in this paper is the adaptation of an existing very deep convnet trained for image classification for the task of saliency prediction.
Previous work~\cite{zeiler2014visualizing} has noted how, in image classification tasks, the model parameters from the lowest levels in the convnets converge in a few epochs.
This observation, together with visualization of the filters learned at these layers~\cite{simonyan2013deep}, suggest that these layers perform low-level visual task in vision, such as the detection of colors or textures.
Our hypothesis is that these lower layers trained for classification can also be transferred for the task of saliency prediction.
We propose a second convnet which adapts these pre-trained filters and combines them with new layers specifically trained for saliency.

\subsection{Architecture}

\begin{figure}
	\centering
	\includegraphics[width=\columnwidth]{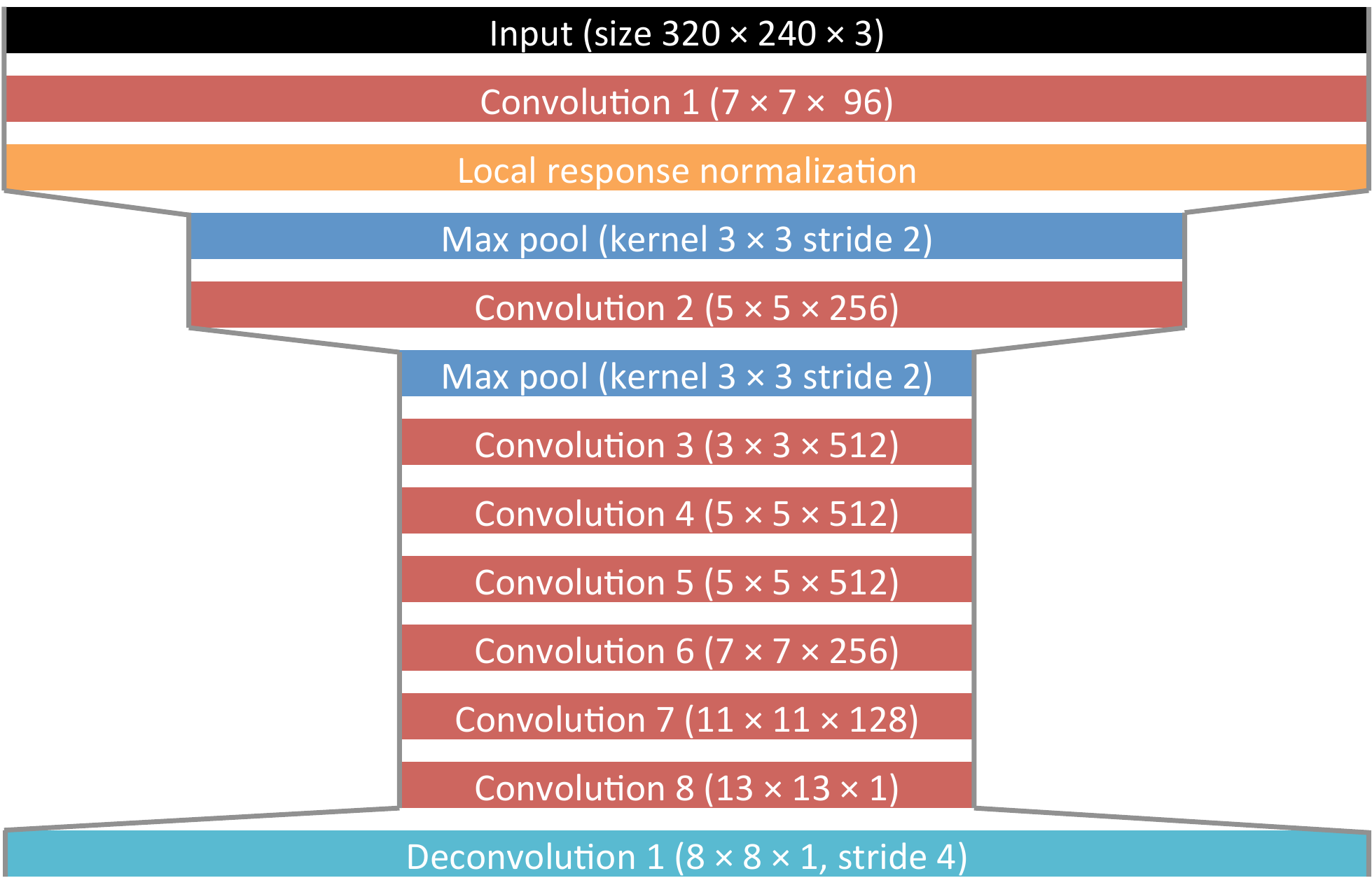}
	\caption{Architecture of the deep convolutional network}
	\label{fig:salnet-arch}
\end{figure}

Figure~\ref{fig:salnet-arch} illustrates the layer architecture of the network, composed of 10 weight layers and a total of 25.8 million parameters.
The architecture of the first 3 weight layers is compatible with that of the VGG network from~\cite{chatfield2014devil}. Each convolutional layer is followed by a rectified linear unit non-linearity (ReLU). Pooling layers follow the first two convolutional layers, effectively reducing the width and height of the feature maps in the intermediate layers by a factor of four. A deconvolution layer follows the final convolution to produce a saliency map that matches the input width and height.

To choose the final network architectures, we experimented with many different different variants, testing each on a held-out validation set of 1,000 images. In general we found that: 1) adding more layers improves accuracy; 2) adding more feature maps per layer usually improves accuracy too; and 3) using dropout regularization did not significantly improve accuracy but did increase training time. The final network design was primarily constrained in resolution, number of layers, and layer depth by the amount of available GPU memory.

We used transfer learning to initialize the weights for the first three convolutional layers with the pre-trained weights from the \emph{VGG\_CNN\_M} network from~\cite{chatfield2014devil}. This acts as a regularizer and improves the final network result. The remaining weights were initialized randomly using the strategy from~\cite{he2015rectifiers}.

\subsection{Training}

We trained our network on 9,000 of the 10,000 training images in the SALICON dataset, setting aside 1,000 images for validation (ground truth for the SALICON validation set had not yet been released when this network was first trained). We used several standard pre-processing techniques on both the input images and the target saliency maps. We subtracted the mean pixel value of the training set from the image pixels to zero center them and rescaled the resulting values linearly to be in the interval $[-1,1]$. We similarly preprocessed the saliency maps by subtracting the mean and scaling to $[-1,1]$. Both the input images and the saliency maps were downsampled by half to $320 \times 240$ prior to training.

The network was trained using stochastic gradient descent with Euclidean loss using a batch size of 2 images for 24,000 iterations. During training, the network was validated against the validation set after every 100 iterations to monitor convergence and overfitting. We used the standard $L^2$ weight regularizer (weight decay), and halved the learning rate every 100 iterations. The network took approximately 15 hours to train on a NVIDIA GTX Titan GPU running the Caffe framework \cite{jia2014caffe}. We normalized the base learning rate by the number of predictions per image, to give a learning rate of $0.01 / (320 \times 240) \approx 1.3 \times 10^{-7}$. Using a larger learning rate causes the learning to diverge.

The network was trained on inputs of size $320 \times 240$, but in principle, it can handle images of any size, since it only consists of convolutional and pooling layers. In practice, the input size is constrained by the amount of GPU memory (or RAM) needed to store the outputs of the intermediary layers. Nevertheless, the network has the advantage that it can be sized to match the aspect ratio of any image, and indeed use this approach for the images in the MIT300 benchmark in the next section.

%% file: 5_results.tex
\section{Experiments}
\label{sec:experiments}


\subsection{Memory requirements}

The architectures of the two networks present different requirements in terms of memory resources.
These resources are dedicated to two different tasks: the parameters that define the network, and the blob data that characterizes network response at the different processing stages.

The parameters that define the network are fit during training, and, together with the architecture layout, correspond to the actual characterization of the network.
These parameters characterize the output of each neuron in the net, which can be defined as $f(w^T x+b)$, where $w$ describes the filter parameters in the convolutional layers, $b$ corresponds to the biases and $f$ is the non-linearity.
Each neuron, therefore, has parameters $w$ and $b$, which are fit during backpropagation.

The data associated to the input image is the second source of memory requirements.
The input image is hierarchically process in the convnet, creating multiple intermediate feature maps (or data blobs) after each processing stage.

Table \ref{tab:memory} presents the complementary memory requirements for each of the two convnets.
These values have been obtained from the architectures of the shallow and very deep networks described in Figures \ref{fig:juntingnet} and \ref{fig:salnet-arch}, respectively.
The estimation assumes 32-bit floating points to store parameters and layer output (4 bytes per value).
The memory estimate for blob data assumes test time (forward pass only):
at train time this value is doubled to account for the error signal during backpropagation.

The number of parameters for both networks are much lower than the very deep networks used in classification. For example, the 19 layers version of VGG net requires 144 million parameters \cite{Simonyan14c}.

Our shallow network requires far less memory for the layer outputs, but has significantly more parameters (due to the fully connected layers). This explains why our deep network does not overfit, whereas stronger regularization is necessary to fit the shallow one. Since the shallow network needs less memory for the layer outputs, it is possible to make batch size on this network very large at test time, allowing it to process many more images at once.


\begin{table}
\begin{center}
\begin{tabularx}{\columnwidth}{Xrr}
\toprule
				& Shallow   		& Deep \\
\midrule
Data 			& 2.29 MB 			& 123.65 MB \\
Parameters  	& 244.64 MB			&  98.44 MB \\
\cmidrule(r){2-3}
Total (train) 	& 249.22 MB			& 345.74 MB	\\
Total (test)	& 246.93 MB			& 222.09 MB \\
\bottomrule
\end{tabularx}
\end{center}
\caption{Approximate memory requirements for each convnet.}
\label{tab:memory}
\end{table}

\subsection{Datasets}
\label{ssec:datasets}

The two convnets were assessed using images and metrics considered in the public MIT Saliency Benchmark \cite{judd2009learning} LSUN challenge 2015 \cite{zhanglarge}.

These four datasets capture a broad range of image types and experimental set ups.
The MIT300 and CAT2000 datasets are smaller in size, but provide fixations points captured in a controlled environment of expert users. 
On the other hand, the iSUN and SALICON datasets have a large amount of saliency maps corresponding to images from the existing SUN and MS CoCo dataset, but these maps were collected via crowdsourcing on Amazon Mechanical Turk, exposing them to crowdsourcing loss~\cite{carlier2015assessment}.

\begin{description}
\item[SALICON \cite{jiang2015salicon} ] This is the largest dataset available for saliency prediction and was used to train our models.
It was built from images of the \textit{Microsoft CoCo: Common Objects in Context} \cite{lin2014microsoft} dataset, which inspired the SALICON naming: \textit{SALIency in CONtext}. The pixel-wise semantic annotations provided by CoCo allow combining and comparing saliency data with semantic ones.
However, the saliency maps in SALICON were not collected with eyetrackers as in most popular datasets for saliency prediction, but with mouse clicks captured in a crowdsourcing campaign.
\item[iSUN \cite{xu2015turkergaze}] The iSUN dataset has been built with an online game using  webcams to track player eye gaze. The dataset uses natural scene images from the SUN database \cite{xiao2010sun}, a large dataset  organized in 397 scene categories.
\item[MIT1003 and MIT300 \cite{judd2009learning}] This dataset is the most well-known among saliency prediction researchers. It is accompanied by an online benchmark maintained by its authors.
The MIT1003 dataset consists of both images and fixation points that can be used for training. The fixation points for the MIT300 dataset are not public: the dataset can only be used for benchmarking.
The stimuli images in these datasets consist of indoor and outdoor natural scenes from the Flickr Creative Commons and LabelMe \cite{russell2008labelme} datasets. The MIT datasets are the smallest of the considered datasets, so results on these sets have the most potential for overfitting. 
\end{description}

\begin{table*}
\begin{center}
\begin{tabular}{lllllll}
\toprule
Dataset							&	Description			 & Capture device 	& Observers & Train 	& Validation & Test \\
\midrule
SALICON	\cite{jiang2015salicon} & Microsoft CoCo \cite{lin2014microsoft} & Mouse clicks	& Crowd & 10,000 	& 5,000	& 5,000 \\
iSUN 	\cite{xu2015turkergaze} & SUN \cite{xiao2010sun} & Eyetracker	& Crowd & 6,000 	& 926	& 2,000 \\
MIT300 \cite{mit-saliency-benchmark}	& Flickr and LabelMe \cite{russell2008labelme} 	& Eyetracker	& 39 & - & -	& 300 \\
\bottomrule
\end{tabular}
\end{center}
\caption{Description of the three datasets used in our experiments.}
\label{tab:datasets}
\end{table*}

\subsection{Results}

Saliency prediction evaluation has received the attention of several researchers, resulting in various proposed approaches.
Our experiments consider several of these, in a similar way to the MIT saliency benchmark~\cite{Judd_2012}. 
Some of these metrics compare the predicted saliency maps with the maps generated from the fixation points of the ground truth, while some other metrics directly compare with the fixation points.
In the result tables that follow, we have sorted the different techniques based on the AUC Judd metric.

Where not otherwise stated, our convnets were trained with images from the SALICON~\cite{jiang2015salicon} dataset and tested on images from iSUN~\cite{xu2015turkergaze} and MIT300 datasets to avoid overfitting.
The one exception to this is our submission for the LSUN 2015 challenge, where our shallow network was trained with training and validation data from iSUN, and assessed on the test partition.

The presented shallow and deep convnets were compared quantitatively on the validation partition of the iSUN dataset.
The results (Table~\ref{tab:iSUN-val}) show a similar performance of both networks in the the 926 images of this dataset.

\begin{figure*}
		\begin{center}
		\includegraphics[width=0.8\linewidth]{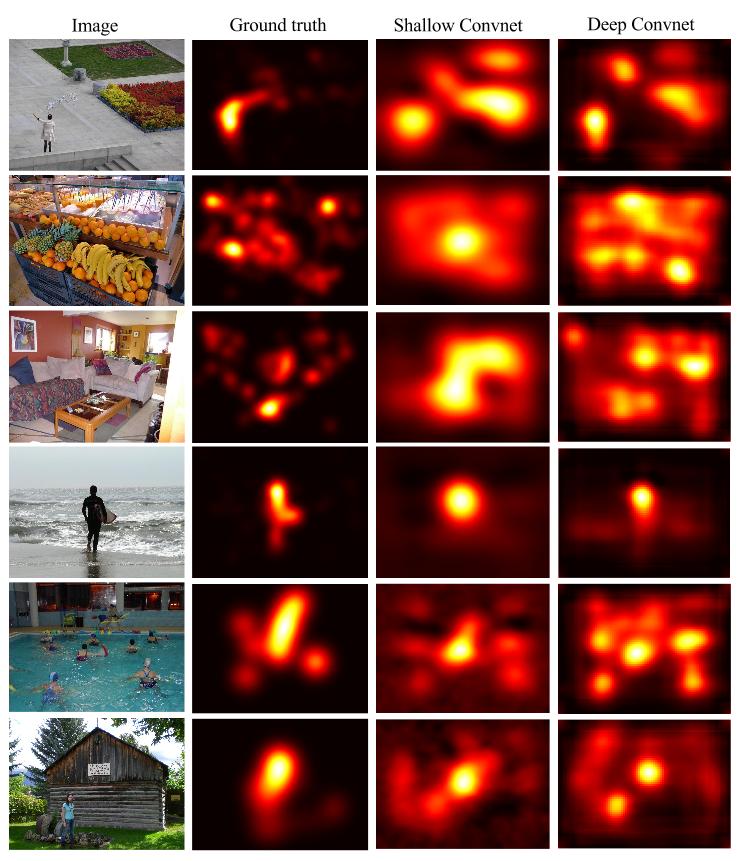}
        \end{center}
		\caption{Saliency maps generated by our shallow and deep network on the SALICON and iSUN validation data.}
		\label{fig:qualitative}

\end{figure*}

Figure~\ref{fig:qualitative} presents a qualitative comparison of the two networks, showing the predicted saliency maps alongside the ground truth fixation maps.
These examples show a different behaviour between the two networks, with the shallow one presenting a bias towards the central part of the image.
The deep network, on the other hand, offers a higher spatial resolution thanks to its architecture with larger feature maps.

\begin{table}
\begin{center}
\begin{tabularx}{\columnwidth}{Xrrr}
\toprule
	AUC				 			& Shuffled 	& Borji &  Judd \\
\midrule
Deep Convnet 					& 0.63 			& 0.78			& 0.80 \\
Shallow Convnet	 				& 0.64 			& 0.77			& 0.79 \\
\bottomrule
\end{tabularx}
\end{center}
\caption{Comparison of AUC measures for our deep and shallow convnets on iSUN validation.}
\label{tab:iSUN-val}
\end{table}



Our shallow convnet was the winner of the 2015 LSUN saliency prediction challenge~\cite{zhanglarge}. 
This challenge required participants to evaluate their algorithms on the test partitions of the iSUN and the SALICON datasets.
Our network was trained only with images from the training and validation partitions of each dataset separately, so images from different datasets were never mixed for these experiments.
Table~\ref{tab:iSUN} and Table~\ref{tab:SALICON} include the results provided by the organizers of the challenge for the iSUN and SALICON datasets.
The scores obtained for every measure considered demonstrate the superior performance of our shallow network compared with the other participants.

\begin{table*}
\begin{center}
\begin{tabular}{lrrrrr}
\toprule
					&	Similarity 	& CC 		& AUC shuffled 	& AUC Borji & AUC Judd \\
\midrule
\textbf{Shallow Convnet (iSUN)}	& \textbf{0.6833} 		& \textbf{0.8230}	& \textbf{0.6650} 		& \textbf{0.8463}	& \textbf{0.8693} \\
Xidian  				& $0.5713$ 		& $0.6167$	& $0.6484$ 		& $0.7949$	& $0.8207$ \\
WHU IIP 			& $0.5593$ 		& $0.6263$	& $0.6307$ 		& $0.7960$	& $0.8197$ \\
LCYLab 				& $0.5474$ 		& $0.5699$	& $0.6259$ 		& $0.7921$	& $0.8133$ \\
Rare 2012 Improved \cite{riche2013rare2012}	& $0.5199$ 		& $0.5199$	& $0.6283$ 		& $0.7582$	& $0.7846$ \\
\midrule
Baseline: BMS \cite{zhang2013saliency}		& $0.5026$ 		& $0.3465$	& $0.5885$ 		& $0.6560$	& $0.6914$ \\
Baseline: GBVS \cite{harel2006graph}		& $0.4798$ 		& $0.5087$	& $0.6208$ 		& $0.7913$	& $0.8115$ \\
Baseline: Itti \cite{itti1998model}		& $0.4251$ 		& $0.3728$	& $0.6024$ 		& $0.7262$	& $0.7489$ \\
\bottomrule
\end{tabular}
\end{center}
\caption{Results for the iSUN test set, according to the LSUN Challenge 2015.}
\label{tab:iSUN}
\end{table*}

\begin{table*}
\begin{center}
\begin{tabular}{lrrrrr}
\toprule
					&	Similarity 	& CC 		& AUC shuffled 	& AUC Borji & AUC Judd \\
\midrule
\textbf{Shallow Convnet}	& \textbf{0.5198} 		& \textbf{0.5957}	& \textbf{0.6698} 		& \textbf{0.8291}	& \textbf{0.8364} \\
WHU IIP 			& $0.4908$ 		& $0.4569$	& $0.6064$ 		& $0.7759$	& $0.7923$ \\
Rare 2012 Improved \cite{riche2013rare2012}	& $0.5017$ 		& $0.5108$	& $0.6644$ 		& $0.8047$	& $0.8148$ \\
Xidian			 	& $0.4617$ 		& $0.4811$	& $0.6809$ 		& $0.7990$	& $0.8051$ \\
\midrule
Baseline: BMS \cite{zhang2013saliency}		& $0.4542$ 		& $0.4268$	& $0.6935$ 		& $0.7699$	& $0.7899$ \\
Baseline: GBVS \cite{harel2006graph}		& $0.4460$ 		& $0.4212$	& $0.6303$ 		& $0.7816$	& $0.7899$ \\
Baseline: Itti \cite{itti1998model}		& $0.3777$ 		& $0.2046$	& $0.6101$ 		& $0.6603$	& $0.6669$ \\
\bottomrule
\end{tabular}
\end{center}
\caption{Results for the SALICON test set, according to the LSUN Challenge 2015. }
\label{tab:SALICON}
\end{table*}

Both of the proposed convnets were also evaluated on the MIT300 dataset~\cite{judd2009learning} of the MIT Saliency Benchmark~\cite{Judd_2012}.
Table~\ref{tab:mit300} compares our results with some other top performers in this benchmark. 
Our deep convnet achieves similar results to the ones obtained by Deep Gaze 1~\cite{kummerer2014deep} at the upper part of the table. The shallow convnet performs worse but still in the upper part of a table which, in its full version, compares 47 different models.




\begin{table*}
\begin{center}
\begin{tabular}{lrrrrr}
\toprule
									&	Similarity 	& CC 		& AUC shuffled 	& AUC Borji & AUC Judd \\
\midrule
Baseline: Infinite Humans 			& $1.00$ 		& $1.00$	& $0.80$ 		& $0.87$	& $0.91$ \\
SALICON  \cite{huan2015salicon} (*)	& $0.60$ 		& $0.74$	& $0.74$ 		& $0.85$	& $0.87$ \\
DeepFix \cite{srinivas2015deepfix} (**)	& $0.67$ 	& $0.78$	& $0.71$ 		& $0.80$	& $0.87$ \\
Deep Gaze 1 \cite{kummerer2014deep}	& $0.39$	& $0.48$ 	& $0.66$		& $0.83$	& $0.84$  \\
\textbf{Deep Convnet}	& \textbf{0.52} 	& \textbf{0.58}	& \textbf{0.69} & \textbf{0.82}	& \textbf{0.83} \\
BMS \cite{zhang2013saliency} 	& $0.51$ 		& $0.55$	& $0.65$ 		& $0.82$	& $0.83$ \\

eDN \cite{vig2014large} 		& $0.41$ 		& $0.45$	& $0.62$ 		& $0.81$	& $0.82$ \\
GBVS \cite{harel2006graph}		& $0.48$ 		& $0.48$	& $0.63$ 		& $0.80$	& $0.81$ \\
Judd \cite{judd2009learning} 	& $0.42$ 		& $0.47$	& $0.60$ 		& $0.80$	& $0.81$ \\
\textbf{Shallow Convnet}		& \textbf{0.46} & \textbf{0.53} & \textbf{0.64} & \textbf{0.78}	& \textbf{0.80} \\
Mr-CNN \cite{liu2015predicting} & $0.48$ 		& $0.48$	& $0.69$ 		& $0.75$	& $0.79$ \\
Rare 2012 Improved \cite{riche2013rare2012}	& $0.46$ 		& $0.42$	& $0.67$ 		& $0.75$	& $0.77$ \\
Baseline: One human		& $0.38-0.46$ 		& $0.52-0.65$	& $0.63-0.67$ 	& $0.66-0.71$	& $0.80-0.83$ \\
\bottomrule
\end{tabular}
\end{center}
\caption{Results of the MIT300 dataset. (*-to be published, (**-non-peer reviewed}
\label{tab:mit300}
\end{table*}

A detailed analysis of the MIT300 results suggests a potential dataset bias~\cite{torralba2011unbiased} in these benchmarks.
Notice how while GBVS \cite{harel2006graph} clearly outperforms our shallow convnet for MIT300 (Table \ref{tab:mit300}), its results are much lower than our shallow convnet or Rare 2012 Improved \cite{riche2013rare2012} for the iSUN (Table \ref{tab:iSUN}) and SALICON datasets (Table \ref{tab:SALICON}).
Unlike many of the other top performing results on the MIT benchmark (DeepFix~\cite{srinivas2015deepfix}, Deep Gaze 1~\cite{kummerer2014deep}, eDN~\cite{vig2014large}, and Judd~\cite{judd2009learning}), our networks were not trained or fine-tuned on the MIT1003 dataset, but trained purely on SALICON data. Our strong results across multiple datasets and benchmarks demonstrate the generality of our models.

%% file: 6_conclusions.tex
\section{Conclusions}
\label{sec:conclusions}

We propose a novel end-to-end approach for training convnets in the task of saliency prediction.
The excellent results of both architectures in state-of-the-art benchmarks demonstrate the superior performance of our convnets with respect to hand-crafted solutions and highlight the importance of an end-to-end formulation of saliency prediction.

The comparison between our shallow and deep networks trained on SALICON data has provided similar results for the iSUN dataset, but a better result for the deep network on MIT300.
On the other hand, the shallow network requires less memory at train time and generates saliency maps much faster because it has fewer layers. 
Both networks rank highly in the MIT300 benchmark despite not being trained on this dataset.
This clearly demonstrates the generalization performance of the networks and robustness to dataset biases.

%% file: 9_acknowledgements.tex
\section*{Acknowledgements}
\label{sec:acknowledgements}

This publication has emanated from research conducted with the financial support of Science Foundation Ireland (SFI) under grant number SFI/12/RC/2289.
This work has been developed in the framework of the project BigGraph TEC2013-43935-R, funded by the Spanish Ministerio de Econom\'ia y Competitividad and the European Regional Development Fund (ERDF). 
The Image Processing Group at the UPC is a SGR14 Consolidated Research Group recognized and sponsored by the Catalan Government (Generalitat de Catalunya) through its  AGAUR office.
We gratefully acknowledge the support of NVIDIA Corporation with the donation of the GeForce GTX Titan Z used in this work.